%% file: main.tex
\documentclass[10pt,twocolumn,letterpaper]{article}
\usepackage{cvpr}

\usepackage{times}
\usepackage{epsfig}
\usepackage{graphicx}
\usepackage{amsmath}
\usepackage{amssymb}

\usepackage{multirow}
\usepackage{colortbl}
\usepackage{arydshln} 
\usepackage{booktabs}
\usepackage{subcaption}
\usepackage{multirow}
\usepackage{nicematrix} 
\usepackage{tablefootnote}
\definecolor{citecolor}{HTML}{0071bc}
\usepackage[pagebackref=false,breaklinks=true,letterpaper=true,colorlinks,citecolor=citecolor,bookmarks=false]{hyperref}
\usepackage{pifont}
\usepackage{tikz}
\usepackage[T1]{fontenc}
\newlength\savewidth\newcommand\shline{\noalign{\global\savewidth\arrayrulewidth
  \global\arrayrulewidth 1pt}\hline\noalign{\global\arrayrulewidth\savewidth}}
\newcommand{\tablestyle}[2]{\setlength{\tabcolsep}{#1}\renewcommand{\arraystretch}{#2}\centering\footnotesize}
\renewcommand{\paragraph}[1]{\vspace{1.25mm}\noindent\textbf{#1}}

\newcolumntype{x}[1]{>{\centering\arraybackslash}p{#1pt}}
\newcolumntype{y}[1]{>{\raggedright\arraybackslash}p{#1pt}}
\newcolumntype{z}[1]{>{\raggedleft\arraybackslash}p{#1pt}}

\newcommand{\app}{\raise.17ex\hbox{$\scriptstyle\sim$}}

\definecolor{deemph}{gray}{0.6}

\definecolor{baselinecolor}{gray}{.9}
\newcommand{\baseline}[1]{\cellcolor{baselinecolor}{#1}}
\definecolor{zeroshotcolor}{gray}{.3}

\definecolor{LightCyan}{rgb}{0.92,1,1}

\def \ModelName {\textit{\tldrcolor{TL;DR}}\xspace}
\def \CCThree          {\textit{CC3M}\xspace}
\def \ModelNameCCThree        {\textit{\ModelName-CC3M}\xspace}
\def \CCTweleve          {\textit{CC12M}\xspace}
\def \ModelNameCCTweleve        {\textit{\ModelName-CC12M}\xspace}
\def \YFCC         {\textit{YFCC15M}\xspace}
\def \ModelNameYFCC        {\textit{\ModelName-YFCC15M}\xspace}
\def \LAION          {\textit{LAION40M(128)}\xspace}
\def \ModelNameLAION        {\textit{\ModelName-LAION40M(128)}\xspace}

\definecolor{demphcolor}{RGB}{144,144,144}
\newcommand{\demph}[1]{\textcolor{demphcolor}{#1}}

\definecolor{LightCyan}{rgb}{0.92,1,1}
\definecolor{darkergreen}{RGB}{21, 152, 56}
\definecolor{red2}{RGB}{252, 54, 65}

\definecolor{bluebell}{rgb}{0.84, 0.84, 0.92}
\newcommand*\colorcmark[1]{%
  \expandafter\newcommand\csname #1cmark\endcsname{\textcolor{#1}{\ding{51}}}%
}
\colorcmark{green}

\newcommand*\colorxmark[1]{%
  \expandafter\newcommand\csname #1xmark\endcsname{\textcolor{#1}{\ding{55}}}%
}

\definecolor{Highlight}{HTML}{39b54a}  

\newcommand{\hl}[1]{\textcolor{Highlight}{#1}}

\colorxmark{red}

\definecolor{titlecolor}{rgb}{0.6,0.4,0.8}
\newcommand\tldrcolor[1]{{\textcolor{titlecolor}{#1}}}

\begin{document}

\title{{\LARGE \tldrcolor{$\mathcal{T}$}}oo {\LARGE \tldrcolor{$\mathcal{L}$}}arge\tldrcolor{\huge $\mathcal{;}$} {\LARGE \tldrcolor{$\mathcal{D}$}}ata {\LARGE \tldrcolor{$\mathcal{R}$}}eduction for Vision-Language Pre-Training}

\author{Alex Jinpeng Wang \quad Kevin Qinghong Lin \quad David Junhao Zhang\quad \\ Stan Weixian Lei \quad Mike Zheng Shou$\thanks{Corresponding Author.}$\\
[3pt]
Show Lab, National University of Singapore\quad \\}
\maketitle

\begin{abstract}
This paper examines the problems of severe image-text misalignment  and  high redundancy in the widely-used large-scale Vision-Language Pre-Training (VLP) datasets. 
To address these issues, we propose an efficient and straightforward Vision-Language learning algorithm called \ModelName, which aims to compress the existing large VLP data into a small, high-quality set.  
Our approach consists of two major steps. 
First,  a codebook-based encoder-decoder  captioner is developed to select  representative samples. 
Second, a new caption is generated to complement the original captions for selected samples, mitigating the text-image misalignment problem while maintaining uniqueness.
As the result, \ModelName enables us to reduce the large dataset into a small set of high-quality data, which can serve as an alternative pre-training dataset. 
This algorithm significantly speeds up the time-consuming pretraining process.
Specifically, \ModelName can compress the mainstream VLP datasets at a high ratio, e.g., reduce well-cleaned CC3M dataset from 2.82M to 0.67M ($\sim$24\%) and noisy YFCC15M from 15M to 2.5M ($\sim$16.7\%).
Extensive experiments with three popular VLP models over seven downstream tasks show that VLP model trained on the compressed dataset provided by \ModelName can perform similar or even better results compared with training on the full-scale dataset.
The code is available at \url{https://github.com/showlab/datacentric.vlp}.
\end{abstract}

\section{Introduction}

\input{figures/1-Motivation.tex}

The recent “scale-is-everything” viewpoint has become a widely accepted notion in the Vision-language Pre-training (VLP) communtity~\cite{cc3m,cc12m,CLIP,align,flamingo}.
According to this view, the scale of the data has increased from the original tens of thousands-level (e.g., COCO~\cite{coco} and VG~\cite{VisualGenome}) to millions-level (e.g., CC3M~\cite{cc3m} and CC12M~\cite{cc12m}), and even up to billions-level (e.g., YFCC100M~\cite{thomee2016yfcc100m}, WIT400M~\cite{CLIP}, and LAION400M~\cite{schuhmann2021laion}). 
Approaches \cite{coca,CLIP,align} trained on these large-scale data show remarking performance improvement in various downstream tasks.

However, simply scaling-up data brings two critical challenges:
\emph{i.} Larger image-text datasets lead to  more training cost (e.g., Pretraining CoCa takes about 5 days on 2,048 CloudTPUv4 chips~\cite{coca}) and storage overhead, which is difficult to afford. \emph{ii}. Obtaining high-quality VLP data requires massive data and well-designed collecting/filtering pipeline, which is expensive. For instance, the CC3M~\cite{cc3m} data was obtained after filtering  5 billion collected images. These challenges are daunting and may impede the participation of numerous researchers in the VLP community.

\input{tables/compress_compare}

In this study, we stop hunting for larger-scale data blindly and ask an important question: \textit{Does employing a larger dataset always result in better performance in VLP?}
To explore and answer this question, we begin with a simple experiment. 
First, we utilize a pre-trained BLIP~\cite{blip} model to calculate the Image-Text Matching (ITM) scores for all  samples in the clean CC3M dataset. 
Subsequently, we remove a portion of the samples with the lowest ITM scores and evaluate the transfer learning results, as shown in Figure \ref{fig:1_motivation}.
Surprisingly, discarding 50\% of the samples slightly improves performance.
This remarkable finding challenges the prevailing belief that employing larger amounts of data invariably leads to superior VLP outcomes. 

This experiment suggests removing certain data points can actually improve the model's ability to learn and generalize. 
Moreover, considering the  performance improvements after removing the low ITM score data, we can infer the existence of significant  misalignment  between the textual and visual modalities in many text-image data pairs (see Figure~\ref{fig:4_score_visualization} and the supplementary material for more evidences).
These discoveries present promising potentiality to enhance the performance of models that depend on a smaller volume of VLP data.

Driven by above analysis and recent advance in dataset pruning~\cite{sorscher2022beyond}, we present a simple, effective and scalable algorithm called \ModelName that aims to improve data efficiency for visual-language pretraining. The \ModelName  has a powerful codebook-based captioner, which contains a visual encoder, a look-up codebook and a text decoder. Here is how it works: First, \ModelName feeds each image into the visual encoder and determines the corresponding codes of the image by measuring the similarity between the codebook and the embedding generated by the encoder. Given a large pool of image-text pairs, \ModelName clusters the samples based on their image corresponding codes and selects a representative subset of samples from each cluster. Then, \ModelName further refines the caption of the selected samples via text decoder to reduce text-image misalignment. 
By doing so, \ModelName is able to significantly reduce the size of the training dataset while maintaining the high quality.

In this work, we employ \ModelName on widely-used CC3M, CC12M, YFCC100M and LAION400M datasets and evaluate small size data on three widely-used frameworks including CLIP~\cite{CLIP}, ViLT~\cite{vilt}, and BLIP~\cite{blip} for data efficiency pretraining with seven representative visual-language downstream tasks. 
The results show that, with only $10\% - 25\%$ data obtained by \ModelName, frameworks achieve similar or even better performance compared with the full-scale dataset. 
We hope our findings can inspire the community to reconsider data efficiency for VLP rather than blindly utilizing increasingly massive datasets.

\section{Related Work}

\subsection{Data-Efficient Learning}
Recent successes in deep learning are largely attributed to the vast amount of data~\cite{imagenet,CLIP}.
However, collecting massive amounts of data is expensive and raises concerns about privacy and copyright~\cite{yu2023dataset}. 
As a result, the research community has become increasingly interested in data-efficient learning, which includes:

\textbf{Dataset Distillation}~\cite{datasetdistillation,zhao2020dataset,kai2022cafe}  compress a large dataset into a small set of synthetic samples, enabling models trained on the smaller dataset to achieve competitive performance with those trained on the original dataset.
However, these techniques are only effective on relatively small datasets at low resolutions, such as CIFAR~\cite{cifar}, and their performance deteriorates significantly when applied to larger-scale datasets.
For example, the accuracy of a model trained on the state-of-the-art MMT's generated data is only 33.8\% on the ImageNet-1K~\cite{imagenet} test result~\cite{mtt}, while pre-training on real ImageNet-1K achieves over 80\% accuracy~\cite{cui2022scaling}.
Furthermore, these methods necessitate supervised class labels, which are not suitable for multimodal data.

\textbf{Data Pruning}~\cite{toneva2018empirical,paul2021deep} assumes high redundancy in large datasets, selecting only a subset of challenging samples.
\cite{mahmood2022much,paul2021deep} observed that during the entire training process, some examples are learned early and never forgotten, while others can be repeatedly learned and forgotten.  
The related work~\cite{sorscher2022beyond} uses a hard sample selection method to select 80\% samples of the ImageNet dataset, and the model trained on selected samples approximating training on all data.
Another recent work, CiT~\cite{xu2023cit}, also proposes to train models with dynamic training data.

\textbf{Neural Data Server} (NDS)~\cite{neuraldataserver,cao2021scalable,lin2022sept} proposes a large-scale search engine to identify the most useful transfer learning data from large corpus. 
While these methods can be extended to multi-modality data, a similar idea has also been applied in NLP~\cite{nlpfromscratch}. 
However, this setting assumes that the user has access to all downstream data and needs to train the downstream task using additional retrieval data.

In this work, we are different from previous techniques in that we attempt to compress large-scale multi-modal data for the first time, leading to comparable performance between the compressed and original vision-language datasets.
We provide a comparison of our approach with these related works in Table~\ref{tabs:dataset_compress_compare}.
\input{figures/3-Main_PPL.tex}

\subsection{Visual-Language Pre-training}
Large-scale Vision-Language Pre-training (VLP) involves training on extensive multi-modality data and evaluating performance on various downstream vision-language tasks. Conventional frameworks include the dual-stream architecture~\cite{CLIP}, the one-stream architecture~\cite{vilt, oscar}, and the encoder-decoder architecture~\cite{blip}. Previous works have relied on high-quality, human-annotated datasets such as COCO~\cite{coco} (110K images) and Visual Genome~\cite{VisualGenome} (100K). As model sizes continue to increase, pre-training requires even more data than before~\cite{coco, align, git}, resulting in an extremely high computational cost. However, obtaining large and high-quality multi-modality data is challenging due to the difficulties in annotation. 
In this paper, we aim to democratize VLP research by proposing a general compression method for existing VLP data.

\section{Method}

Our \ModelName is a simple yet effective approach for compressing the Vision-Language Pre-training dataset, leading to further reduction of the training cost.
Our approach consists of two stages: (1) codebook-based captioner training and (2) data reduction including samples selection and caption refining. Figure~\ref{fig:3_main_ppl} illustrates the idea, introduced next.

\subsection{Codebook-based Captioner}
The captioner consists of a visual encoder, a codebook and a text decoder. 
The visual encoder is employed to extract image features. 
Inspired by vector quantisation technique~\cite{vqgan,vq_vae}, we try to quantize the image feature for further clustering  by utilizing a learnable codebook. 
Codebook comprises $K$ learnable embedding vectors, each of which can be regarded as a code. Each token of image features conducts a nearest neighbor look-up  from codebook and finds its corresponding code. 
In this way, image features are quantized into a couple of codes (quantized vectors).  
The quantized vectors are then sent into a text decoder, generating a caption. 
In order to enhance the quality of text generation, we initialize the codebook with the text embedding of $K$ most frequently occurring keywords/keyphrases in the entire dataset, which enables the codebook to contain meaningful and intuitively understandable semantics. 

To train the whole captioner, we utilize  a Language Modeling loss~\cite{bert}, which maximizes the likelihood of the text in an autoregressive manner,  and a  symmetric commitment loss~\cite{vqgan}, which is  specifically designed for codebook. We initially train this captioner on noisy source data and subsequently fine-tune it on smaller-scale datasets, such as COCO~\cite{coco} and VisualGenome~\cite{VisualGenome}. 

\subsection{Data Reduction}

Currently, large-scale datasets exist with serious redundancy~\cite{sorscher2022beyond}. Meanwhile, a large part of texts is noisy and misaligned with images in VLP data. 
See Figure \ref{fig:3_main_ppl} for the example (the caption ``You need to think twice before buying a pet as present'' does not match the image). To overcome these limitations,  we use the learned codebook to condense large-scale noisy data and the learned captioner to reduce the misalignment over image-text pairs.

\input{tables/all_ablation}

\textbf{Samples selection.}
For an encoded image feature with $L$ tokens, we compute an index vector with length $L$.  Each value is the index of the code, which is the closest to each token. This vector maps the features from image space to semantic space so that it reduces the complexity of the image, benefiting and accelerating the cluster process.  
Subsequently, each image sample in the dataset is equipped with an index vector according to the above process and we cluster these vectors into $N$ clusters with  K-Means ( speed up by Faiss~\cite{faiss}). Then we uniformly sample $M\%$ data points from each cluster, producing a small subset of the dataset. We examine various sampling methods and observe that uniform sampling is stable across different scales.

\textbf{Caption refining.}
To alleviate the misalignment problem, we want to improve the text quality using the generated caption.  Generated text $T_g$ is from the text decoder, which takes the quantized vector of the image as input. We simply concatenate $T_g$ with original text  $T_o$  together, denoted as $T = T_o + T_g$, to refine and preserve the original caption's uniqueness while maintaining data diversity.

The compressed small-scale dataset with refined captions is recorded as dataset $D_c$.
At last, we train VLP models on this high quality dataset $D_c$ and expect the model to achieve comparable performance with original full-scale dataset $D$ on downstream Vision-Language tasks.

\textbf{Discussion.}
Considering the serious misalignment problem, it seems quite straightforward to use pure generated high-quality caption $T_g$ to replace original noisy text.
Driven by this idea, we try to pretrain BLIP~\cite{blip} models with $T_o$, $T_g$ and $T_o+T_g$ independently and show the train curve of Image-Text Contrastive (ITC) loss in Figure~\ref{fig:4_train_curve}.
However, we find the model trained with $T_g$ fails into model collapse~\cite{robinson2021can}.
This phenomenon can be explained by captioning collapse~\cite{vinyals2015show,wang2020overview} and one-to-many problem~\cite{young2014image} in image cpationing. 
That is,  the trained captioner will generate fixed or similar captions for different images, which limits diversity in the output and easily leads to trivial solutions for contrastive loss.
On the contrary, the ITC loss for both $T_o$ and $T_o+T_g$ works well and the $T_o+T_g$ converges better.
We also observe the loss of $T_g$ is smaller than other two variants at epoch 0-2, which indicates the generated caption matches well with the image.
Note that this simple stitching operation on caption does not bring additional computation cost for VLP as the max length in text-encoder Bert~\cite{bert} model keeps unchanged for all setting.

\subsection{Technical Details.}
Our \ModelName can be implemented efficiently, and importantly, does not require any large auxiliary model.
The codebook size $K$ is 3000 as default.
The selection of keywords/phrases is implemented using the NLTK \footnote{https://github.com/nltk/nltk}.
We adopt ViT-B/16~\cite{vit} as image encoder and BertLMHead Model~\cite{bert} as text decoder.
The image transformer is initialized from ViT pre-trained on ImageNet~\cite{imagenet}, and the text transformer is initialized from BERTbase~\cite{bert} (BLIP,CLIP) and DistilBERT~\cite{distilbert} (ViLT).
In this way, the token length $L$ is 196 as default.
The cross-attention is computed over image embedding and text embedding.
To show the generality of compressed dataset, we test $D_c$ on three different and representative VLP architectures: dual-stream CLIP~\cite{CLIP}, one-stream ViLT~\cite{vilt} and Fusion-encoder Blip~\cite{blip} on various downstream tasks.
All these models are trained under the same setting with different datasets.
\input{figures/4_train_curve}

\input{tables/ft_and_zero_shot_retrieval}
\input{tables/vqa_nlvr_refcoco}

\section{CC3M Experiments}
We first study dataset reduction on well-cleaned \textit{CC3M}~\cite{cc3m} which heavily filters web crawled pairs and only keeps 0.1\% of the raw data.  
This dataset contains a total of 2.8 million images.
We employ our \ModelName to compress the CC3M dataset, then conduct pre-training and fine-tuning evaluations on both  original and compressed datasets.
Following our ablation study, we transfer the pre-trained model to seven Vision-Language tasks downstream and fine-tune it through end-to-end training to evaluate its performance.

\paragraph{Training.}
We utilize PyTorch~\cite{pytorch} to implement our models and trained them on 8 NVIDIA A100 GPUs to reduce the data samples. 
For Vision-Language Pre-training, we utilize 2 nodes, each equipped with 16 GPUs.
The model is pre-trained for 20 epochs with a batch size of 1260 and an AdamW~\cite{adamw} optimizer with a weight decay of 0.05. 
During training, we apply a learning rate warm-up to 3e-4 and a linear decay with a rate of 0.85.
For image augmentation, we utilize RandAugment \cite{randaugment} and apply all of the original policies except  color inversion. 
This decision is based on the recognition of the crucial role that color information plays in the data.
For pre-training, images were randomly cropped to a resolution of $224 \times 224$. 
We then increase this to $384 \times 384$ for fine-tuning downstream tasks. 
Further information about the training hyperparameters for downstream tasks can be found in the supplementary material.

\subsection{Main Properties}
\label{sec:ablation}
We ablate our \ModelName using the default setting in Table~\ref{tab:ablations} (see caption). 
Several intriguing properties are observed.

\paragraph{Module deconstruction.}
In Table \ref{tab:compoent_ablation} we analyze the impact of different components in \ModelName.
We establish a baseline by randomly selecting 25\% of the data from CC3M (first row). 
Our results show that codebook-based sampling outperforms random selection by 3.2\% in TR@1. 
We also observe that both \textit{codebook-based sampling} and \textit{caption refinement} are crucial and the combination of them achieves optimal downstream performance.

\paragraph{Sample selection.}
In Table \ref{tab:selection_strategy} we study the sample selection strategy in Stage 2.
We sample 25\% data in each cluster by default.
For \textit{Gradient-based}, we train a tiny network to conduct VLP pretrained with ITC~\cite{oscar}, ITM~\cite{oscar} and LM~\cite{bert}. 
Then we select samples  which contribute most to the gradients in each cluster.
\textit{Large distance:}
Another perspective is that data points on the border of each cluster are more important than those at the center~\cite{bora2014comparative}.
So we first compute the center of each cluster and then choose the sample that has the largest distance from the center of each cluster.
We also report the result of \textit{hard-sample} selection from~\cite{sorscher2022beyond}.
We observe that all these variants produce similar results except \textit{large distances}.
This suggests that the clustering step, rather than the selection step, plays a key role in data compression during Stage 2.
To maintain simplicity, we choose uniform sampling as the default method.

\input{tables/msrvtt_retrieval}

\paragraph{Codebook initialization.}
In Table~\ref{tab:codebook_selection} we compare different initialization strategies.
The xavier means all parameters in the codebook are initialized with xavier initialization~\cite{xavier}.
For the object tags initialization, following previous works~\cite{butd,vinvl}, we use the 1600 object tags from Visual Genome~\cite{VisualGenome} and extract text feature with a pre-trained BERT~\cite{bert}.
With same training setting, the keywords achieve a 0.8\% TR@1 improvement and a 0.7
\% IR@1 improvement over xavier. This result is expected as the text embeddings provide contextual information and simplify the learning process.

\input{tables/zero_shot_image_classification}

\paragraph{Codebook vs. Image embedding.}
In Table~\ref{tab:codebook_comparison}, we investigate different ways of cluster sampling. 
First, we remove the codebook from Stage-1 and use image embedding instead. 
Alternatively, we directly cluster images using the image embedding~\cite{blip} of images from BLIP model (pre-trained on 200M Image-text pairs). 
We observe the image embedding leads to much better result than text embedding.
This is reasonable because clustering visual-similarity samples with text only is difficult.
We observe that clustering depended on our codebook performs better than both image embedding and text embedding. 
This demonstrates that our codebook can efficiently project image embedding to semantic space, benefiting cluster process.  

\paragraph{Cluster sampling ratio.}
Table~\ref{tab:sample_ratio} varies the sampling ratio of each cluster from 10\% to 100\%. 
We are surprised to find that a low sampling ratio can still produce effective results. 
With only 25\% of the data and the \ModelName model, we are able to achieve a 1.9\% improvement on TR@1 and a 0.8\% improvement on IR@1 over the full-scale baseline. 
Additionally, we observe that larger sampling ratios lead to even better results. 
Since our focus is on \textit{achieving similar transfer learning results with fewer samples}, we use a default sampling ratio of 25\% to minimize computation costs.

\paragraph{Cluster numbers.}
In Table~\ref{tab:cluster_number}, we investigate the impact of cluster number on Stage 2 by increasing it from 300 to 30K. 
We observe that using more clusters results in a slight improvement at the beginning and becomes stable when the number of clusters exceeds 3K. 
Moreover, all results consistently outperform the random selection baseline. 
Therefore, we use 3K clusters as the default in this work, as it performs well on fine-tuning tasks.

\subsection{Transfer Learning Experiments.}
We conduct an extensive evaluation of transfer learning in downstream tasks using the model pre-trained on our compressed \ModelNameCCThree and source \CCThree with 3 architectures. 
Our evaluation primarily focuses on the core tasks of three categories that examine: (1) cross-modality alignment, (2) image captioning and multi-modality understanding capabilities, and (3) visual recognition.
The baseline in this section is the model trained on CC3M dataset.

\subsubsection{Cross-modality Alignment Task}
\paragraph{Image-Text retrieval.}
Fine-grained world region alignment plays a critical role in this task.
We report both image-to-text retrieval (TR) and text-to-image retrieval (IR) on the COCO~\cite{coco} and Flickr30K~\cite{flickr30k} benchmarks. 
For the BLIP~\cite{blip} model, we adopt an additional re-ranking strategy, following the original implementation. 
In Table \ref{tab:ft_and_zero_shot_retrieval}, we also report zero-shot retrieval results. 
We found that \ModelName achieves comparable results with the baselines on all metrics and surprisingly performs quite well on zero-shot results. 
For example, for the BLIP~\cite{blip} architecture, our method leads to a 6.4\%  improvement (from 42.3\% to 48.7\%) in Recall@1 of image-to-text retrieval on MSCOCO. 
All results suggest that a small part of refined image-text pairs is enough to learn good alignment.

\paragraph{Zero-shot video retrieval.}
In this experiment, we analyze the generalization ability of our method to video-language tasks. 
Specifically, we perform zero-shot transfer to text-to-video retrieval and evaluate the models trained on COCO-retrieval in Table \ref{tab:zsl_video_retrieval}. 
We uniformly sample 8 frames from each video to process the video input and concatenate the frame features into a single sequence.
These models trained on our compressed dataset outperform the baseline on all metrics, demonstrating the generality of \ModelName.

\input{figures/4-2-CaptionerVisualization}

\input{figures/4-1-ClusterVisualization.tex}

\subsubsection{Image Captioning and Multi-modality Understanding Tasks}

\paragraph{Image captioning.}
The task involves describing an input image, which we evaluate using No-Caps and COCO datasets. 
Both datasets are fine-tuned on COCO with the Language Modeling (LM) loss.
We adopt a zero-shot setting for No-Caps dataset, and start each caption with the phrase ``a picture of'' for the BLIP architecture. 
We do not pre-train using COCO to avoid information leakage. 
Our results outperform baseline with a much smaller quantity of pre-training data, as shown in Table \ref{tab:vqa_nlvr}.

\paragraph{Visual question answering (VQA).}
We evaluate our model's performance on the VQA task \cite{vqa}, where the model needs to provide an answer based on an image and a question. 
We consider it as an answer generation task that allows open-vocabulary VQA for better results, following previous works \cite{albef,blip}. 
The results are presented in Table \ref{tab:vqa_nlvr}. 
The BLIP trained on \ModelNameCCThree outperforms baseline by 1.4\% on test-dev splits, demonstrating the effectiveness of our compressed dataset for improving VQA performance.

\input{figures/4-3-GeneratedSamples}

\paragraph{Visual reasoning.}
The Natural Language Visual Reasoning (NLVR$^2$) \cite{nlvr2} task is a binary classification task that requires the model to reason about two images and a question in natural language. 
Multi-modal reasoning is crucial for the completion of this task.
We observe that BLIP trained on our dataset achieved 78.0\% accuracy compared to 76.2\% achieved by the CC3M, as shown in Table \ref{tab:vqa_nlvr}.

\paragraph{Cross-modality grounding.}
Referring Expression (RE) Comprehension requires the model to select the target object from a set of image regions proposals, based on the query description. 
This task heavily relies on visual-grounding ability. 
The models are evaluated on ground-truth objects, and we evaluate RE Comprehension on RefCOCO+~\cite{refcoco}. 
The results are reported in Table \ref{tab:vqa_nlvr}, and we observe that \ModelNameCCThree achieves better results.

\subsubsection{Visual Recognition Tasks}
Besides the cross-modality task, we also explore a uni-modality task, mainly image classification. 
Specifically, we fix the image encoder and explore zero-shot image classification. 
Since the vision encoder is loaded from a pretrained model, this task demonstrates the impact of training a Vision-Language model with noisy image-text pairs. Specifically, it shows how such training affects the well-learned representation derived from a human-crafted dataset.
We show the results in Table \ref{tabs:zs-image} and observe noisy data leads to significant Catastrophic Forgetting.
For example, the CLIP model drops down to only 58.3 accuracy with noisy data training.
We also observe our \ModelName shows steady improvement for all architectures over random selection. 
Unfortunately, the classification performance for \ModelNameCCThree is slightly worse than the full-scale CC3M for the CLIP and BLIP architectures. 
Both of these architectures have independent image encoders like ViT to extract image embeddings. 
This indicates that this task heavily relies on visual diversity, which is different from multi-modal tasks, and our method reduces the visual diversity potentially. 
For the ViLT model, this architecture adopts a shared backbone for both visual and text, and we observe the slightly different results. 
We guess that multi-modality interaction in early-fusion affects the classification result.

\input{tables/generated_sample_comparison}

\subsection{Visualization}

\paragraph{Generated caption visualization.}
We show the generated caption in Figure~\ref{fig:4_generated_caption}.
It is evident that the original captions can be highly abstract and difficult to match  their respective images, even for human observers sometimes. 
For instance, when the ITM score is as low as 0.04, matching the figure with its corresponding caption becomes arduous. 
Such challenging cases can potentially harm the cross-modality alignment.
In contrast, we observe that the generated captions describe the image very well and sometimes offer helpful complementary information.
For example, ``bus'' and ``castle'' in the middle example.

\input{figures/score_visualization}

\input{tables/other_datasets}

\paragraph{Codebook-based cluster visualization.}
Figure \ref{fig:4_visualization} displays the codebook grouping result achieved with simple K-Means.
Clusters are sets of data points with similar characteristics, often defined by their features or attributes.
Interestingly, we observe that the model cluster samples ``accurate'', meaning that these samples have semantic similarity rather than simple appearance. 
For instance, the model classifies ``dollars'' and ``piggy bank'' together, even though they differ significantly in appearance.

\subsection{More Investigation}

\paragraph{Is image generation possible?}
To ease the misalignment problem of image-text pairs, instead of simply selecting representative samples, a potential and naive idea is to generate images from text.
To this end, we randomly sample 0.3M subset of CC3M and generate image from text with three popular text to image generation models, VQ-GAN~\cite{vqgan}, DALLE 2~\cite{dalle2} and Stable Diffusion~\cite{stablediffusion}.
We display the generated samples in Figure \ref{fig:4_generated_sample}.
We observe that the generative models struggle with complex scenarios, but are capable of generating simple prompts like ``dog'' proficiently. 
In addition, generation methods only produce visual cues in a fixed vocabulary, potentially reducing data diversity.

Next, we pre-train BLIP models on these generated data and evaluate it on COCO Retrieval. 
In Table \ref{tabs:generated_compare} we observe the results of transfer learning depend on the quality of generated samples, with those generated by stable diffusion being particularly effective.
However, there still exists a significant gap between the generated data and the real dataset (e.g., 52.4\% vs. 58.3\% on TR@1).
We believe that higher-quality and diverse generated images may lead to comparable results with real images in the near future.


\paragraph{Explore the misalignment problem.}
Figure~\ref{fig:4_score_visualization} shows the Image-text Matching (ITM) score distribution for both \CCThree and our \ModelNameCCThree data (the visualization about more datasets is reported in the supplementary).
We observe a lot of samples of original \CCThree at low matching score even tends to zero, which indicates the current dataset has serious misalignment problems.
Since image-text matching (ITM) loss and image-text contrastive (ITC) loss are used in all architectures, these samples will damage the multimodal representation learning.
When adopting our \ModelName, we observe that the matching score tends to be higher and has very few samples with low ITM score.

\vspace{-1mm}
\section{Transfer to other VLP datasets}

We study data compression performed in two categories shown below: clean data that involves human-based offline filter pipelines and raw data that has not undergone cleaning.
For clean data, in addition to \CCThree, we explore the well-cleaned, high-quality dataset \CCTweleve~\cite{cc12m}.
Then, we study the raw data YFCC100M~\cite{thomee2016yfcc100m} and LAION400M~\cite{schuhmann2021laion}.
\CCTweleve~\cite{cc12m} contains 12 million image-text pairs specifically meant to be used for vision-and-language pre-training.   
These data are collected by relaxing the data collection pipeline as in CC3M.
YFCC15M~\cite{CLIP} is a subset of the multilingual and noisy YFCC100M~\cite{thomee2016yfcc100m} that contains English captions.
LAION400M~\cite{schuhmann2021laion} is a large-scale noisy dataset that provides URLs with captions for download. 
To control the computation cost and reduce the storage overhead, we randomly sample a 40M subset of LAION400M and download images at a resolution of 128 $\times$ 128.
So, we record the dataset as \ModelNameLAION, and the performance over downstream tasks could be improved with higher resolution.
More exploration about video-text datasets is reported in the supplementary material

We use BLIP as the default architecture and evaluate our \ModelName on different datasets and show the results in Table~\ref{tab:other_datasets}.
Surprisingly, with only 2.5M (16.7\%) data, \ModelNameYFCC leads to similar results with 15M raw data over all metrics except Imagenet.
More results with different backbones are reported in the supplementary material.
For \LAION, when using 8M data (20\%), the model trained on our dataset consistently outperforms the baseline method on six downstream tasks.
We noticed that the compression rate of \LAION is less than that of \YFCC.
This may be due to the fact that the collection of \LAION has already been filtered with CLIP similarity, reducing the impact of the misalignment problem.

\section{Conclusion and Discussion}
This paper presents \ModelName, a novel and pioneering algorithm for selecting and generating high-quality image-text pairs from noisy Vision-Language Pre-training (VLP) data, thereby contributing to the field of VLP.
\ModelName incorporates a text generation process into learning to reduce serious misalignment problem. 
Our experiments demonstrate three widely-used architectures leads to comparable results and much smaller training cost when learning from our generated dataset. 
Additionally, we demonstrate that the misalignment problem can be effectively addressed using our simple \ModelName. However, the choice of the highest compression ratio is done manually rather than learned. Furthermore, achieving even higher compression ratios for VLP models remains a challenge, and text-to-image generation models may be helpful in this regard. We hope that this perspective will inspire future research.

{\small
\bibliographystyle{ieee_fullname}
\bibliography{main}
}

\end{document}

%% file: figures/1-Motivation.tex
\begin{figure}[h]
  \centering
    \includegraphics[width=\linewidth]{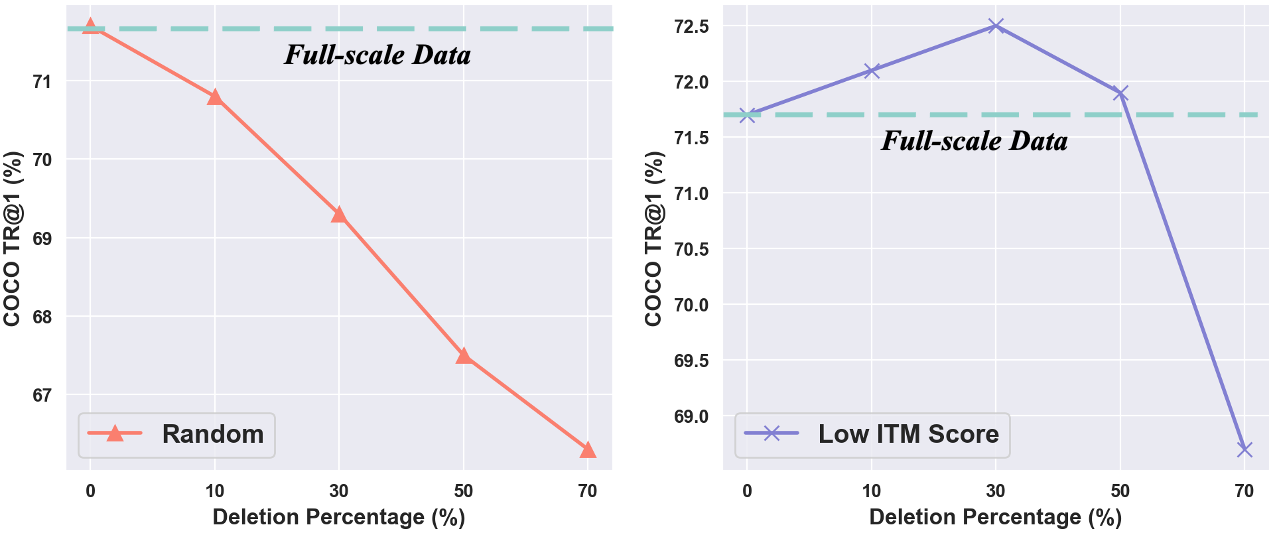}
       \vspace{-.5em}
   \caption{
   \textbf{Does using more data really lead to better performance in VLP?}
   Instead of training on the full-scale CC3M dataset, we delete data with low image-text matching score.
   We find that BLIP~\cite{blip} model pretrained on 50\% reserved data even obtains better result than full-scale dataset on downstream COCO retrieval~\cite{coco}.
   This observation exposes there exists serious \textit{misalignment} between text\&visual modalities and data redundancy in dataset.
   }
   \vspace{-1.2em}
   \label{fig:1_motivation}
\end{figure}

%% file: tables/compress_compare.tex
\begin{table*}[t]
\footnotesize
\centering
\begin{tabular}{@{}llllccll@{}}
    Method & Year & Data Type & Compression Ratio$\uparrow$&Task Agnostic &Large-scale&Supervision&Generation/Selection\\
    \shline
    Dataset Distillation~\cite{datasetdistillation} & 2017& Image &99\%-99.99\% & \redxmark &\redxmark & Class Label & Generation \\
    Data Pruning~\cite{sorscher2022beyond} & 2022& Image & 20\%-30\%   & \redxmark  & \greencmark& Class Label & Selection\\
    Neural Data Server~\cite{neuraldataserver} & 2020 & Multi-modality & 94\%-98\% &\redxmark& \greencmark &Image-text Pairs&Selection \\
    \ModelName (ours)&- & Multi-modality & 75\%-90\%& \greencmark&\greencmark&Image-text Pairs&Generation+Selection \\
\end{tabular}
\vspace{-1em}
\caption{\label{tabs:dataset_compress_compare} 
\textbf{Data-efficient learning methods}. 
"Large-scale" means that the methods are effective when used on datasets that are very large in size. 
The "task agnostic" means that the methods can be used regardless of the specific downstream task, and without any prior exposure to the associated data.
}
\vspace{-1em}
\end{table*}

%% file: figures/3-Main_PPL.tex
\begin{figure*}[h]
  \centering
\includegraphics[width=.95\linewidth]{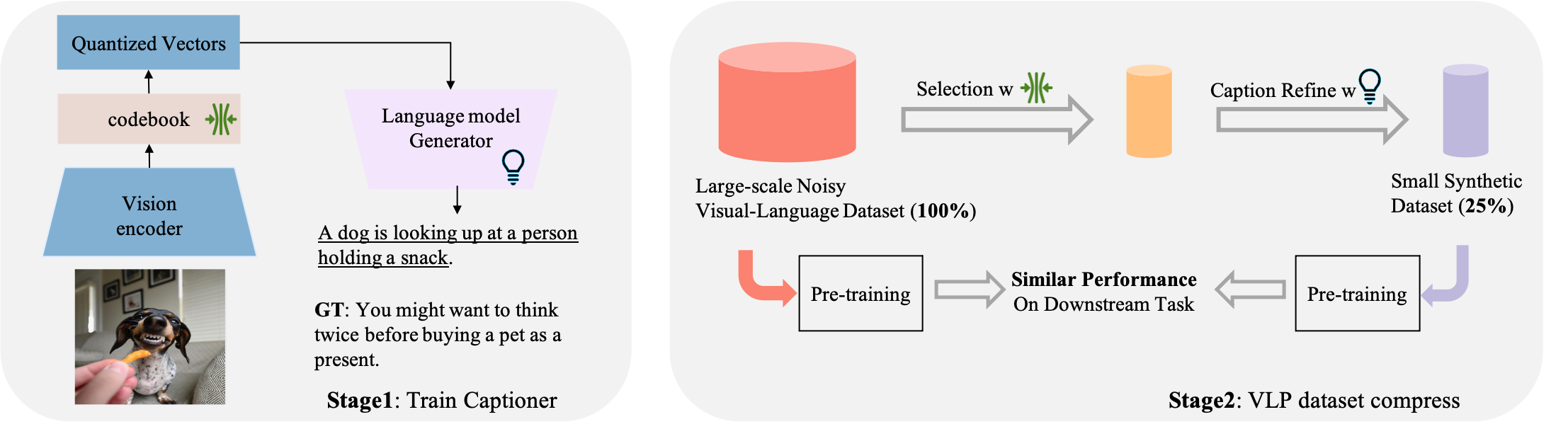}
\vspace{-.5em}
   \caption{
   \textbf{Our \ModelName architecture}.
   We first train a codebook-based captioner in Stage1.
   Then the learned codebook and captioner are used to reduce VLP data in Stage 2.
     Pre-training on the reduced dataset achieves similar performance to the original full-scale dataset across downstream tasks.
   }
   \vspace{-1em}
   \label{fig:3_main_ppl}
\end{figure*}

%% file: tables/all_ablation.tex
\begin{table*}[h]
\centering
\subfloat[
\textbf{Component ablation}.  
Both the sampling and refining operation are important to the downstream retrieval.
\label{tab:compoent_ablation}
]{
\centering
\begin{minipage}{0.29\linewidth}{\begin{center}
\tablestyle{4pt}{1.05}
\begin{tabular}{x{27}x{27}x{20}x{20}}
sampling & refining  & TR@1&  IR@1 \\
\shline
& & 65.3  & 49.8 \\
\checkmark& & 68.5 &51.9\\
&\checkmark& 69.4 &52.3\\
\checkmark&\checkmark & \baseline{\bf 72.8 } & \baseline{\bf 54.8} \\
\end{tabular}
\end{center}}\end{minipage}
}
\hspace{2em}
\subfloat[
\textbf{Sample-selection strategy}.
The different way to select samples.
\label{tab:selection_strategy}
]{
\begin{minipage}{0.29\linewidth}{\begin{center}
\tablestyle{4pt}{1.05}
\begin{tabular}{y{75}x{20}x{20}}
case & TR@1 & IR@1 \\
\shline
gradient-based & 72.9 & \bf 54.8 \\
hard-sample~\cite{sorscher2022beyond} & \textbf{73.1} & 54.5 \\
uniform & \baseline{72.8} & \baseline{\bf 54.8} \\
large distance & 72.3 & 53.1 \\
\end{tabular}
\end{center}}\end{minipage}
}
\hspace{2em}
\subfloat[
\textbf{Codebook Initialization}.
An codebook initialized with keywords is more stable.
\label{tab:codebook_selection}
]{
\begin{minipage}{0.29\linewidth}{\begin{center}
\tablestyle{1pt}{1.05}
\begin{tabular}{y{66}x{24}x{24}}
case & TR@1 & IR@1 \\
\shline
{xavier~\cite{xavier}} & 72.0 & 54.1  \\
{key words/phrases} & \baseline{\textbf{72.8}} & \baseline{\textbf{54.8}} \\
{object tags} & 72.5 & 54.4  \\
\multicolumn{3}{c}{~}\\
\end{tabular}
\end{center}}\end{minipage}
}
\vspace{.3em}
\subfloat[
\textbf{Clustering feature}.  
Codebook is better than Image Embedding at same scale.
\label{tab:codebook_comparison}
]{
\centering
\begin{minipage}{0.29\linewidth}{\begin{center}
\tablestyle{4pt}{1.05}
\begin{tabular}{y{97}x{20}x{20}}
case & TR@1&  IR@1 \\
\shline
Image Embedding & 70.6  & 52.3 \\
Text Embedding& 69.0 & 50.4 \\
BLIP Image Embedding~\cite{blip} & 72.3 & 54.5 \\
Codebook & \baseline{\bf 72.8 } & \baseline{\bf 54.8} \\
&\\
\end{tabular}
\end{center}}\end{minipage}
}
\hspace{2em}
\subfloat[
\textbf{Sampling ratio}.
Sampling 25\% data is enough to beats with full scale.
\label{tab:sample_ratio}
]{
\begin{minipage}{0.29\linewidth}{\begin{center}
\tablestyle{4pt}{1.05}
\begin{tabular}{y{75}x{20}x{20}}
case & TR@1 & IR@1 \\
\shline
\demph{full-scale baseline} & \demph{70.6} & \demph{54.0} \\
10\% & 68.9 & 52.3 \\
25\% & \baseline{72.8} & \baseline{54.8} \\
50\% & 74.8 & 55.2 \\
100\% & \bf 75.1 & \bf 57.7 \\
\end{tabular}
\end{center}}\end{minipage}
}
\hspace{2em}
\subfloat[
\textbf{Cluster Number}.
More clusters not equals better result.
\label{tab:cluster_number}
]{
\begin{minipage}{0.29\linewidth}{\begin{center}
\tablestyle{1pt}{1.05}
\begin{tabular}{y{66}x{24}x{24}}
case & TR@1 & IR@1 \\
\shline
{100} & 71.8 & 53.9  \\
1000 & 72.4 & 54.5 \\
3000 & \baseline{72.8} &\baseline{\textbf{54.8}}  \\
5000 & \bf 72.9 & 54.4 \\
10000 & 72.3 & 54.2 \\
\end{tabular}
\end{center}}\end{minipage}
}
\\
\centering
\caption{\textbf{\ModelName ablation experiments} with BLIP model~\cite{blip} on CC3M. 
We report image-to-text retrieval top-1 (TR@1) and text-to-image retrieval top-1 (IR@1) accuracy (\%) on COCO~\cite{coco} dataset. 
If not specified, the default baseline is: pre-training BLIP model based on ViT-B/16 with 25\% sample of CC3M. 
Default settings are marked in \colorbox{baselinecolor}{gray}.}
\label{tab:ablations} \vspace{-1em}
\end{table*}

%% file: figures/4_train_curve.tex
\begin{figure}[t]
  \centering
    \includegraphics[width=\linewidth]{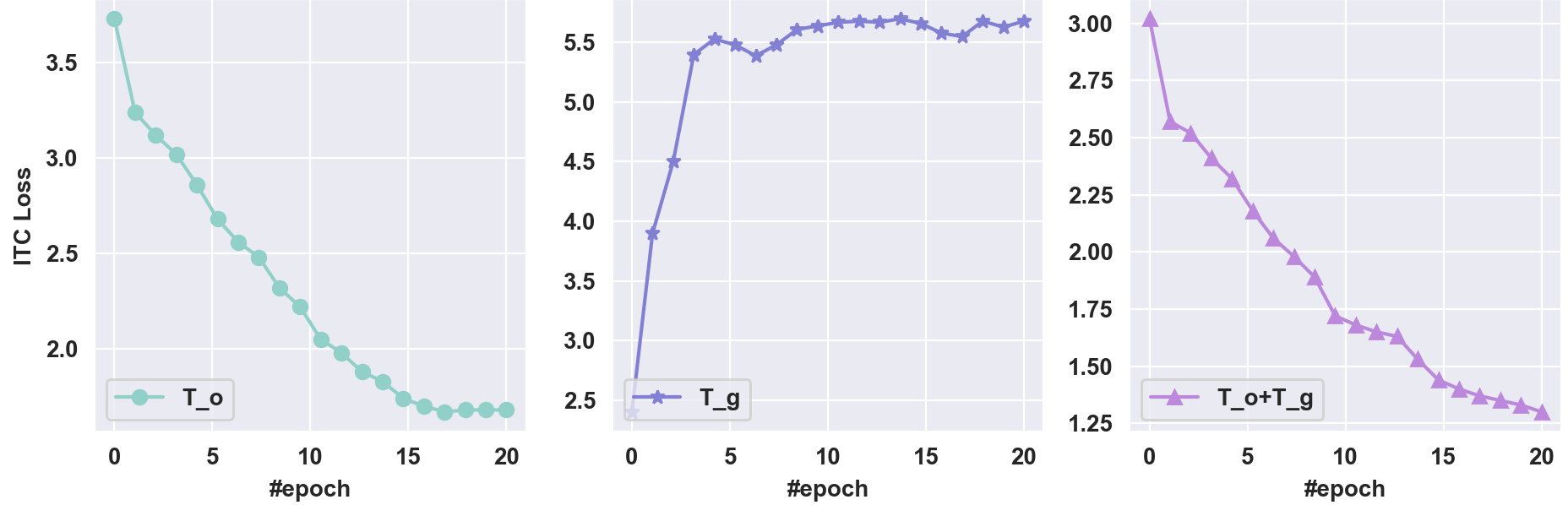}
       \vspace{-1.5em}
   \caption{
   Training curve with CC3M dataset.
     Simply stitching generated text and original text together solved the model collapse problem in Image-text Contrastive Loss.
   }
   \vspace{-1.2em}
   \label{fig:4_train_curve}
\end{figure}

%% file: tables/ft_and_zero_shot_retrieval.tex
\begin{table*}
\centering
\footnotesize
\begin{NiceTabularX}{\linewidth}{y{28}ly{28}|cccccc|cccccc}
\CodeBefore
  \rowlistcolors{4}{=,=,blue!10,=}[restart,,cols={2-15}]Å
\Body
  \footnotesize
                  Method & Dataset & \#Samples  & \multicolumn{6}{c}{ MSCOCO (5K test set)}  & \multicolumn{6}{c}{ Flickr30K (1K test set)}  \\
                  &&&\multicolumn{3}{c}{Image$\rightarrow$ Text}&\multicolumn{3}{c}{Text$\rightarrow$ Image}&\multicolumn{3}{c}{Image$\rightarrow$ Text}&\multicolumn{3}{c}{Text$\rightarrow$ Image}\\ &&&R@1&R@5&R@10&R@1&R@5&R@10&R@1&R@5&R@10&R@1&R@5&R@10\\
\shline
\multirow{4}{*}{CLIP~\cite{CLIP}} & \CCThree~\cite{cc3m} &2.82M  & \bf 60.4 & 85.3 & 93.2 & 48.9 & 75.4 & 84.7 & 77.3 & 91.1 &  \bf 93.2 & 71.6 & 90.1 & 91.4  \\
                  & \ModelNameCCThree  &0.67M  &   60.3 & \bf 85.6 & \bf 93.8 & \bf 49.4 & \bf 77.4 & \bf 86.0 & \bf 82.5 & \bf  91.8 & 92.2 & \bf 72.0 & \bf 90.5 & \bf 92.1  \\
                  & \CCThree ~\cite{cc3m} &2.82M  & 36.2 &  64.3 & 80.1 & 29.9 & 57.9 & 66.9 & 67.4 & 83.2 &  \bf 92.4 & 54.3 & 84.1 &  \bf  90.8  \\
                  & \ModelNameCCThree&0.67M  &   \bf 37.7 &  \bf 64.6 & \bf  80.8 &  \bf 30.7 &  \bf 58.4 & \bf  68.2 & \bf  68.5 &  \bf 85.4 &  92.0 &  \bf 55.6 & \bf  82.1 &  \bf 90.8  \\
\hline
\multirow{4}{*}{ViLT~\cite{vilt}} & \CCThree~\cite{cc3m}&2.82M    &   66.7&\bf 89.2 & 93.8 & 52.5 & \bf 79.3 & 87.1 & 83.8 & 92.0 & 93.2 & 74.0 & 92.0 & \bf 92.8  \\
                  & \ModelNameCCThree&0.67M  &   \bf  67.1&  88.7 & \bf 94.1 &  \bf 53.1 &  78.9 &  \bf 88.2 &  \bf 85.3 &  \bf 92.4 &  \bf 93.6 &  \bf 75.6 &  \bf 92.1 &  92.5 \\
                  & \CCThree ~\cite{cc3m} &2.82M  &   39.2& 68.6 & 77.8 & 30.4 & 53.2 & 66.1 & 70.5 & 88.7 & 92.1 & 57.6 & \bf 84.9 & \bf 92.6 \\
                  & \ModelNameCCThree &0.67M   &   \bf 43.5& \bf 70.8 & \bf 81.4 & \bf 33.9 & \bf 57.9 & \bf 66.8 & \bf 73.2 &  \bf 90.5 &  \bf 93.3 &  \bf  58.6 &   84.7 & 92.4  \\
\hline
\multirow{4}{*}{BLIP~\cite{blip}} & \CCThree ~\cite{cc3m}  &2.82M&   70.9& 91.3 &  \bf 96.1 & 54.3 & 80.2 & 88.0 & 86.3 & 94.1 & 94.8 & 74.8 & 91.6 & 92.6 \\
                  & \ModelNameCCThree &0.67M  &    \bf 72.8&  \bf 91.9 &  95.9 &  \bf 54.8 &  \bf 80.6 &  \bf 89.4 &  \bf 87.5 & \bf  94.8 & \bf  95.3 & \bf  75.7 &  \bf 92.2 &  \bf 93.4  \\
                  & \CCThree ~\cite{cc3m} &2.82M &   42.3& 67.8 & 77.4 & 31.5 & 55.7 & 66.3 & 75.1 & 91.2 & 93.6 & 60.6 & 85.9 & 91.8  \\
                  & \ModelNameCCThree &0.67M &    \bf  48.7  & \bf   73.1 &  \bf 82.7 &  \bf 36.7 &  \bf 60.6 &  \bf 70.4 &  \bf 76.3 &  \bf 91.9 &  \bf 93.9 &  \bf 61.0 &  \bf 87.7 &  \bf 93.0  \\
\end{NiceTabularX}
\vspace{-1em}
\caption{\textbf{Fine-tuning and \colorbox{bluebell}{zero-shot} image-text retrieval} results on MSCOCO and Flickr30K dataset. }
\label{tab:ft_and_zero_shot_retrieval}
\vspace{-.5em}
\end{table*}

%% file: tables/vqa_nlvr_refcoco.tex
\begin{table*}[h]
	\centering	
 \footnotesize
	\begin{tabular}	{ll|lllllllll}
	 \multirow[t]{2}{*}{Dataset}& \multirow[t]{2}{*}{ \#Samples}& \multicolumn{2}{c}{VQA} & \multicolumn{2}{c}{NLVR$^2$} &\multicolumn{3}{c}{RefCOCO+} &\multicolumn{2}{c}{COCO Caption} \\
	  && test-dev & test-std & dev & test-P & val & testA & testB & B@4 & CIDEr \\
	  		\shline 	
    \demph{Random-\CCThree} & \demph{0.67M} & \demph{68.3} & \demph{66.2} & \demph{73.6} & \demph{73.8} &\demph{68.6}&\demph{71.8}&\demph{62.8}&\demph{35.9}&\demph{118.8} \\
	   \CCThree~\cite{cc3m}&2.8M& 71.5 & 71.8 & 76.0 & 76.2 &72.4&76.1&65.3&36.8&121.6 \\
   	    \ModelNameCCThree & 0.67M &73.1$_{\hl{+1.6}}$ & 73.2$_{\hl{+1.4}}$ & 77.7$_{\hl{+1.7}}$ & 78.0$_{\hl{+1.8}}$ &75.1$_{\hl{+2.7}}$&78.5$_{\hl{+2.4}}$&68.4$_{\hl{+3.1}}$&37.6$_{\hl{+0.8}}$&123.8$_{\hl{+2.2}}$ 
	\end{tabular}
    \vspace{-1em}
     	\caption
	{
        \textbf{Comparison with BLIP model pre-trained on different data sources} for VQA, NLVR$^2$, RefCOCO+ and COCO Captioning. 
        ViLT and CLIP architectures can not evaluated on part of these tasks since structural limitations.
	}
	\label{tab:vqa_nlvr}
\end{table*}		

%% file: tables/msrvtt_retrieval.tex
\begin{table}[t]
\centering
\footnotesize

		\begin{tabular}	{y{25}y{50}|cccc}
		Method & {Dataset} &  R@1$\uparrow$ &  R@5$\uparrow$ &  R@10$\uparrow$ & MdR$\downarrow$\\
		\shline
		\multirow{3}{*}{CLIP~\cite{CLIP}} & \demph{Rand-\CCThree} & \demph{15.3} & \demph{34.8} & \demph{46.3} & \demph{13.0}  \\
        &\CCThree~\cite{cc3m} & 19.4 & 37.3 & 47.5 & 11.0\\
        &\ModelNameCCThree& \bf 21.8 &\bf  38.6 & \bf 48.5 & \bf 10.0 \\ 
        \hline
        \multirow{3}{*}{ViLT~\cite{vilt}} &  \demph{Rand-\CCThree} & \demph{18.8} & \demph{38.2} & \demph{49.5} & \demph{11.0}\\
        &\CCThree~\cite{cc3m}& 21.0 &40.5 & 51.5 & 10.0 \\
        &\ModelNameCCThree&\bf 22.5 & \bf 42.7 &\bf 52.4 &\bf 8.0\\
        \hline
        \multirow{3}{*}{BLIP~\cite{blip}} &  \demph{Rand-\CCThree} & \demph{23.3} & \demph{42.8} & \demph{53.3} & \demph{8.0} \\
        &\CCThree~\cite{cc3m}& 26.0 & 46.3 & 58.0 & 7.0\\
        &\ModelNameCCThree&\bf 27.4& \bf 48.7& \bf 59.4 & \bf 6.0\\
		\end{tabular}
    \vspace{-1em}
      \caption
	{
	\textbf{MSRVTT-1K retrieval} using three architectures.
    We created a subset of the \CCThree dataset called \demph{Rand-\CCThree} by randomly selecting the same number of samples as in \ModelNameCCThree.
	}
	\vspace{-2em}
	\label{tab:zsl_video_retrieval}
\end{table}		

%% file: tables/zero_shot_image_classification.tex
\begin{table}[t]
\centering
\footnotesize
\begin{tabular}{@{}y{25}y{47}y{28}|y{18}y{29}y{29}@{}}
    Model &Dataset & \#Samples&  ImNet & ImNet-A & ImNet-R \\
    \shline
    \multirow{3}{*}{CLIP~\cite{CLIP}}& \demph{Rand-\CCThree} &0.67M& \demph{58.3} & \demph{61.8} & \demph{62.3}\\
      &\CCThree~\cite{cc3m}&2.82M& \bf 62.2 & \bf 65.2 & \bf 66.9 \\
     &\ModelNameCCThree&0.67M& 61.4 & 65.0 & 65.7 \\
    \hline
    \multirow{3}{*}{ViLT~\cite{vilt}}&\demph{Rand-\CCThree} &0.67M& \demph{54.3} & \demph{59.8} & \demph{58.4}\\
    &\CCThree~\cite{cc3m}&2.82M&58.6 & 62.9 & \bf 64.2  \\
     &\ModelNameCCThree&0.67M&\bf 59.1& \bf 63.3 & 64.0  \\
    \hline
    \multirow{3}{*}{BLIP~\cite{blip}}&\demph{Rand-\CCThree}&0.67M & \demph{57.3} & \demph{61.8} & \demph{65.2}\\
    &\CCThree~\cite{cc3m}&2.82M& 
    \bf 62.5&\bf 65.5&\bf 68.1\\
    &\ModelNameCCThree&0.67M&62.0&63.9&67.4\\
\end{tabular}
\vspace{-1em}
\caption{\label{tabs:zs-image} \textbf{Zero-shot image classification} results on ImageNet~\cite{imagenet}, ImageNet-A~\cite{imageneta}, ImageNet-R~\cite{imagenetr}.
\textit{There is no free lunch}, as selecting partial samples reduces the visual diversity crucial for classification. 
Despite this, \ModelName still performs significantly better than random selection.
}
\vspace{-1em}
\end{table}

%% file: figures/4-2-CaptionerVisualization.tex
\begin{figure}[t]
  \centering
    \includegraphics[width=\linewidth]{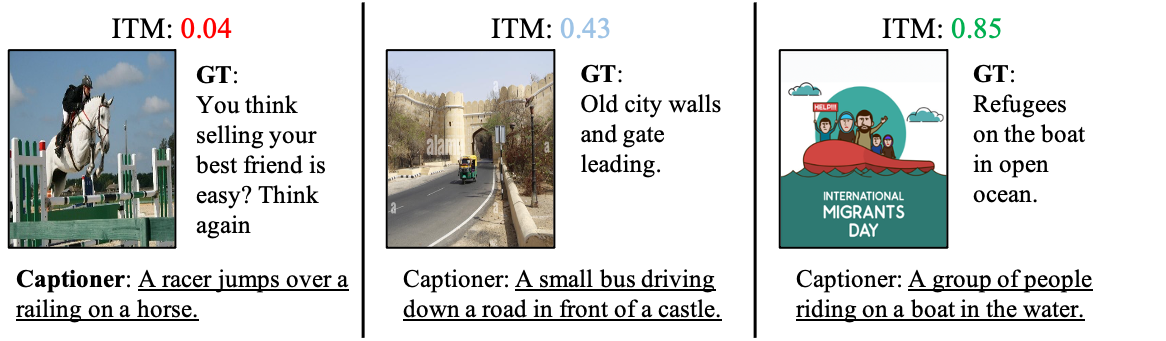}
   \vspace{-2em}
   \caption{
   \textbf{The generated caption} match the image well.
   }
   \vspace{-1em}
   \label{fig:4_generated_caption}
\end{figure}

%% file: figures/4-1-ClusterVisualization.tex
\begin{figure}[h]
  \centering
    \includegraphics[width=\linewidth]{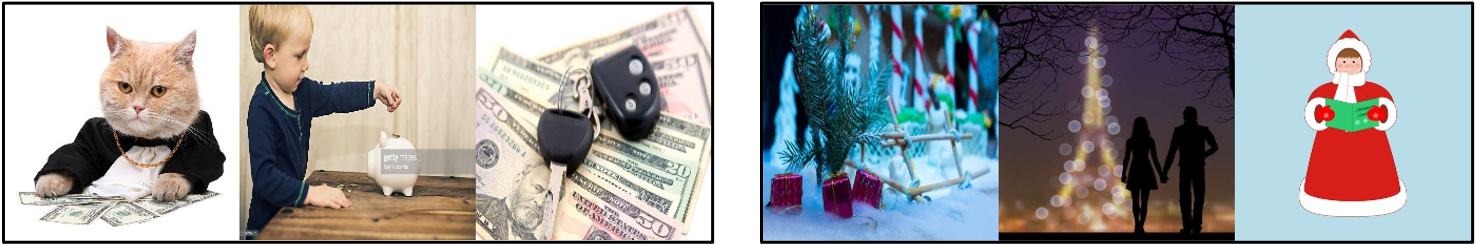}
    \vspace{-1.5em}
   \caption{
   \textbf{The codebook-based clusters visualization.}
    The samples within each cluster exhibit similar contextual characteristics, as opposed to mere visual appearance. 
    For example, the ``Christmas elements'' cluster located at the right.
   }
   \vspace{-1em}
   \label{fig:4_visualization}
\end{figure}

%% file: figures/4-3-GeneratedSamples.tex
\begin{figure}[t]
  \centering
\includegraphics[width=\linewidth]{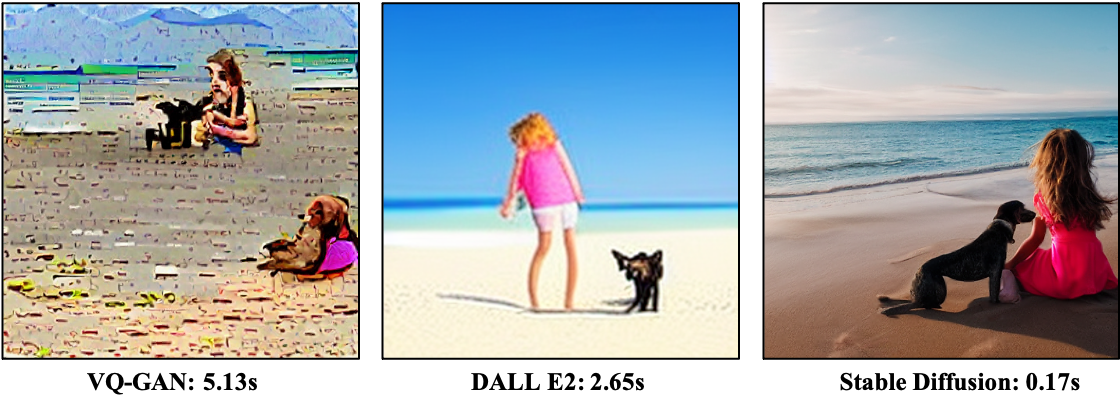}
    \vspace{-1em}
   \caption{
   \textbf{Image generation result} with strong Text-to-image Model.
   The generation time is also reported.
   }
   \vspace{-1em}
   \label{fig:4_generated_sample}
\end{figure}

%% file: tables/generated_sample_comparison.tex
\begin{table}[t]
\footnotesize
\centering
\begin{tabular}{@{}l|cc@{}}
    Method  & TR@1 & IR@1 \\
    \shline
    \demph{real data} & \demph{58.3} & \demph{44.0} \\
    VQ-GAN~\cite{esser2021taming} & 35.2 & 32.4  \\
    DALLE2~\cite{dalle2} (implement from \tablefootnote{https://github.com/LAION-AI/dalle2-laion}) & 44.3 & 38.3 \\
    Stable Diffusion~\cite{stablediffusion} (implement from \tablefootnote{https://github.com/huggingface/diffusers}) & \bf 52.4 & \bf 40.7 \\
\end{tabular}
\vspace{-1em}
\caption{\label{tabs:generated_compare} 
\textbf{Compare different sample generation methods} over 0.3M subset of CC3M.
We first pre-train BLIP model on these generated data and then evaluation on COCO.
}
\vspace{-1em}
\end{table}

%% file: figures/score_visualization.tex
\begin{figure}[t]
  \centering
    \includegraphics[width=.95\linewidth]{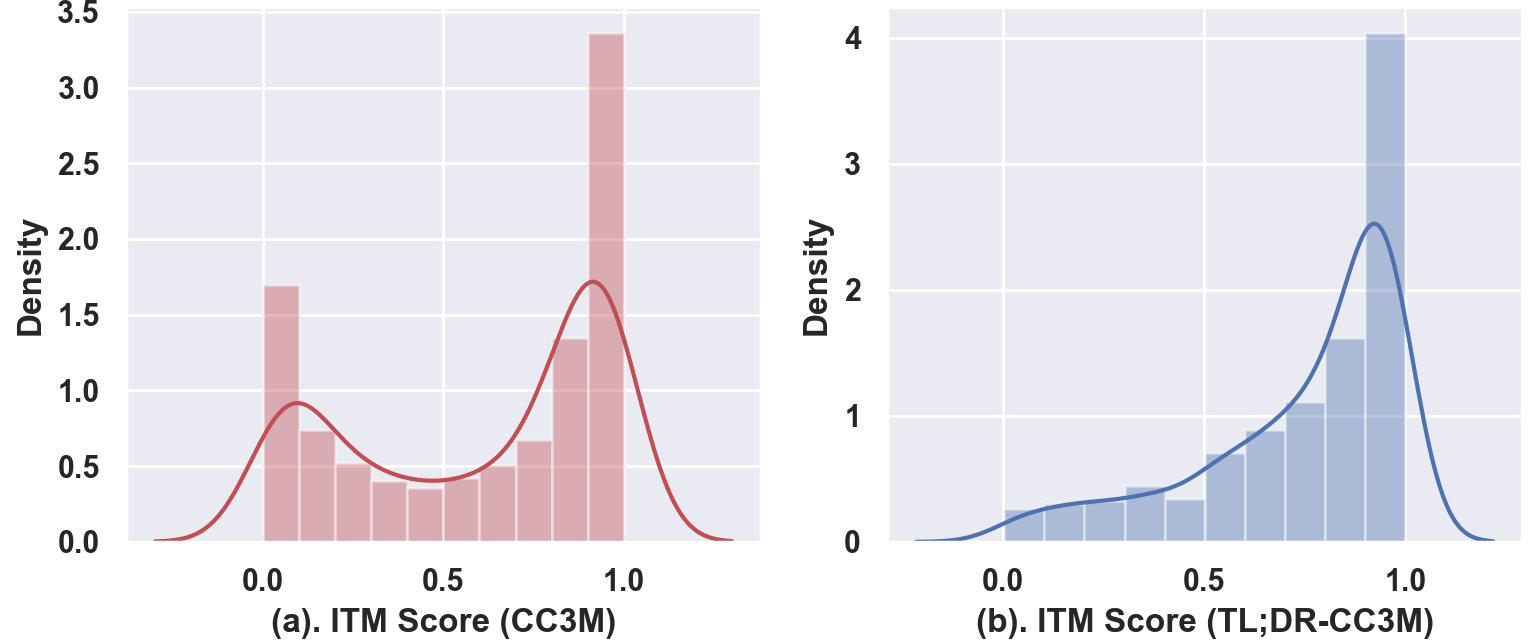}
   \caption{
   \textbf{ITM score distribution.}
   \ModelName alleviates the issue of misalignment in VLP data.
   }
   \vspace{-1em}
   \label{fig:4_score_visualization}
\end{figure}

%% file: tables/other_datasets.tex
\begin{table*}[h]
	\centering	
 \footnotesize
	\begin{tabular}	{y{80}y{20}y{15}|llllllll}
	 \multirow[t]{2}{*}{ Dataset}& \multirow[t]{2}{*}{ \#Samples} & \multirow[t]{2}{*}{ Time} &  VQA &  NLVR$^2$ & 
  RefCOCO & \multicolumn{2}{l}{ Nocaps Captioning} & \multicolumn{2}{l}{Flcikr30K Retrieval} &{ Imagenet} \\
	  &&& test-dev & test-P & val & B@4 & CIDEr & TR@1 & IR@1 & Acc \\
	  		\shline
     \demph{Rand-\CCTweleve}&\demph{2.4M} & \demph{14h} & \demph{71.8} & \demph{76.2} & \demph{72.5} & \demph{36.8} & \demph{121.0} & \demph{82.9} & \demph{73.3} & \demph{61.2} \\
	  \CCTweleve~\cite{cc12m}&10.8M  & 65h& 73.5 & 78.9 & 74.1 & 37.5 & 122.9 &84.7&75.3& 65.3\\
	  \ModelNameCCTweleve & 2.4M  & 14h& 74.1$_{\hl{+0.6}}$ & 78.5$_{-0.4}$ & 74.0$_{-0.1}$ & 38.1$_{\hl{+0.6}}$ & 124.1$_{\hl{+1.2}}$ & 85.5$_{\hl{+0.8}}$ & 76.3$_{\hl{+1.0}}$ & 63.8$_{-1.5}$ \\
   \hline
     \demph{Rand-\YFCC}&\demph{2.5M}  & 15h& \demph{67.2} & \demph{70.5} & \demph{68.1} & \demph{35.2} & \demph{116.3} & \demph{78.8} & \demph{70.5} & \demph{65.4}  \\
	   \YFCC~\cite{thomee2016yfcc100m}&15M & 90h& 70.5& 74.2& 70.6& 35.9 & 118.4 & 81.5 & 72.4 & 67.8 \\
	  \ModelNameYFCC& 2.5M  & 15h& 70.3$_{-0.2}$ &  75.3$_{\hl{+1.1}}$& 72.6$_{\hl{+2.0}}$ & 37.2$_{\hl{+1.3}}$ & 122.5$_{\hl{+4.1}}$ & 82.3$_{\hl{+0.8}}$ & 74.3$_{\hl{+1.9}}$ & 67.3$_{-0.5}$ \\
       \hline
        \demph{Rand-\LAION}&\demph{8M} & \demph{48h} & \demph{70.7} & \demph{75.3} & \demph{73.4} & \demph{34.8} & \demph{113.2} & \demph{80.4} & \demph{72.5} & \demph{68.5} \\
	   \LAION~\cite{schuhmann2021laion} & 40M & 120h & 74.5 & 79.1 & 76.6 & 35.2 & 117.4 & 83.2 & 74.9 & 71.3 \\
   	   \ModelNameLAION& 8M & 48h &  76.3$_{\hl{+1.8}}$ &80.5$_{\hl{+1.4}}$&77.4$_{\hl{+0.8}}$&36.8$_{\hl{+1.6}}$&120.9$_{\hl{+3.5}}$&82.8$_{-0.4}$&76.1$_{\hl{+1.2}}$&70.4$_{-0.9}$  \\
	\end{tabular}
 \vspace{-1em}
     	\caption
	{
		\textbf{Comparison with different source of data on 6 downstream tasks.} 
       BLIP~\cite{blip} is adopted as baseline and (128) means the image resolution is 128$\times$128. We also list the pre-training time, which can be significantly reduced via \ModelName.
	}
    \vspace{-1em}
	\label{tab:other_datasets}
\end{table*}		